\documentclass[10pt,twocolumn,letterpaper]{article}

\usepackage[pagenumbers]{cvpr} % To force page numbers, e.g. for an arXiv version
\usepackage{booktabs}   % For professional-looking tables (\toprule, \midrule, \bottomrule)
\usepackage{pifont}     % For the checkmark and cross symbols (\ding)
\usepackage[table]{xcolor} % For coloring text and table rows (\textcolor, \rowcolor)
\usepackage{soul}
\definecolor{cvprblue}{rgb}{0.21,0.49,0.74}
\usepackage[pagebackref,breaklinks,colorlinks,allcolors=cvprblue]{hyperref}
\usepackage[T1]{fontenc}
\usepackage{booktabs}
\usepackage{makecell}
\usepackage{tabularray}
\usepackage{tabularx}
\usepackage{enumitem}
\usepackage{multirow}
\usepackage{adjustbox}
\usepackage[most]{tcolorbox}
\newcommand{\dataset}{{MyEgo}}
\newcommand{\gemini}{{Gemini 2.5 Pro}}

\newcommand{\cmmnt}[1]{}
\newcommand{\cmark}{\textcolor{green!60!black}{\ding{51}}}%
\newcommand{\xmark}{\textcolor{red}{\ding{55}}}%
\newtcolorbox{promptbox}{
  colback=black!5,   % 背景颜色 (5% 的黑色，即浅灰)
  colframe=black!75, % 边框颜色
  boxrule=0.5pt,     % 边框粗细
  arc=1mm,           % 圆角半径
  boxsep=5pt,        % 文本和边框的间距
}
\def\paperID{***} % *** Enter the Paper ID here
\def\confName{CVPR}
\def\confYear{2026}

\title{

Ego-Grounding for Personalized Question-Answering in Egocentric Videos}

\author{
Junbin Xiao\textsuperscript{1,2}\thanks{Equal Contribution.},
\quad Shenglang Zhang\textsuperscript{1,2*},
\quad Pengxiang Zhu\textsuperscript{2},  
\quad Angela Yao \textsuperscript{2} \\
\textsuperscript{1} University of Science and Technology of China, \textsuperscript{2} National University of Singapore \\
{\tt\small{junbinxiao@ustc.edu.cn, zsl142857@mail.ustc.edu.cn, angela.yao@nus.edu.sg}}
}

\begin{document}
\maketitle

\begin{abstract}
We present the first systematic analysis of multimodal large language models (MLLMs) in personalized question-answering requiring ego-grounding - the ability to understand the camera-wearer in egocentric videos. To this end, we introduce MyEgo, the first egocentric VideoQA dataset designed to evaluate MLLMs' ability to understand, remember, and reason about the camera wearer. MyEgo~comprises 541 long videos and 5K personalized questions asking about ``my things'', ``my activities'', and ``my past''. Benchmarking reveals that competitive MLLMs across variants, including open-source vs. proprietary, thinking vs. non-thinking, small vs. large scales all struggle on MyEgo. Top closed- and open-source models (\eg, GPT-5 and Qwen3-VL) achieve only~46\% and 36\% accuracy, trailing human performance by near 40\% and 50\% respectively. Surprisingly, neither explicit reasoning nor model scaling yield consistent improvements. Models improve when relevant evidence is explicitly provided, but gains drop over time, indicating limitations in tracking and remembering ``me'' and ``my past''. These findings collectively highlight the crucial role of ego-grounding and long-range memory in enabling personalized QA in egocentric videos. We hope MyEgo and our analyses catalyze further progress in these areas for egocentric personalized assistance.
\end{abstract}    
\section{Introduction}
\label{sec:intro}

\begin{figure}[!t]
\centering
\includegraphics[width=0.48\textwidth]{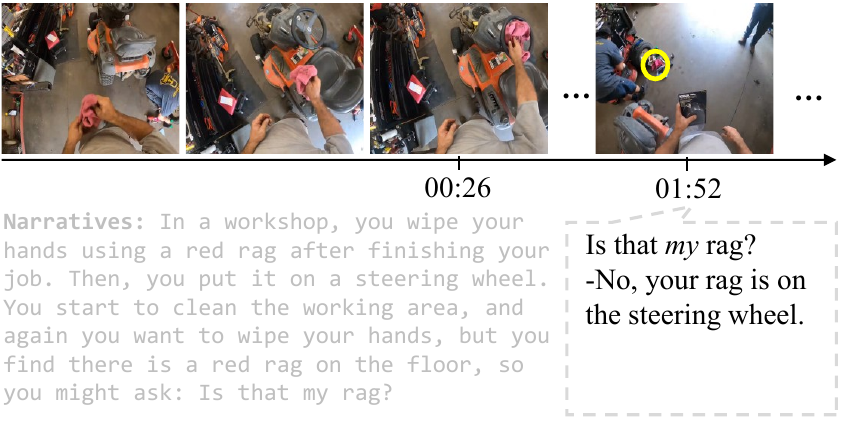}
\caption{Example of egocentric personalized QA concerning \textit{my} demands. The narratives are provided for better understanding.}
\label{fig:intro}
\vspace{-3mm}
\end{figure}

As cameras in smart glasses and other wearable devices become ubiquitous, egocentric videos are emerging as a powerful medium for capturing first-person visual experiences~\cite{damen2018scaling,grauman2022ego4d,yang2025egolife,rossetto2025castle,yan2025teleego,li2025building}. These continuous streams enable personalized assistance and timeline construction, helping users recall what they saw, did or interacted with throughout the day. Achieving this requires \emph{ego-grounding}: understanding “me”, “my things”, “my activities”, and “my past” in ``my'' first-person videos. 

We introduce the notion of ego-grounding as a core prerequisite for personalized egocentric QA assistants. While related ideas appear in 
co-reference resolution \cite{kurita2023refego,sun2025visual}, and personalized VLM \cite{cohen2022my,alaluf2024myvlm}, none addresses the visual and temporal challenges of grounding first-person references (“I”, “my”) in egocentric videos.
Current multimodal large language models (MLLMs) \cite{hurst2024gpt,comanici2025gemini,bai2025qwen2,wang2025internvl3,zhang2024video,yu2025minicpm} appear promising, due to their strong visual reasoning and long-context capabilities.
However, it remains unclear whether current models can actually perform ego-grounding reliably. In egocentric video, the camera-wearer is often only partially visible, with hands, arms, ego-motion or brief reflections, making the task especially challenging.

For example, consider Fig.~\ref{fig:intro}; answering the simple question about “my rag”.  Humans do such tasks easily, yet current MLLMs often fail in such scenarios, revealing that ego-grounding poses challenges not captured by popular VideoQA and egocentric benchmarks \cite{grauman2022ego4d, cheng2024egothink, yang2025egolife}. This example highlights the broader demands of ego-grounding: models must separate the camera-wearer from nearby people, track “my things’’ through visually ambiguous scenes, and recall interactions that may no longer be visible.

Successful ego-grounding therefore requires both spatial discrimination and long-range temporal reasoning. Since these capabilities remain open challenges for state-of-the-art MLLMs \cite{Qwen3-VL,wang2025internvl3, li2025videochat, comanici2025gemini, hurst2024gpt}, we investigate how well current models can perform ego-grounding for answering personalized questions in practice. To that end, we introduce \dataset, the first personalized egocentric VideoQA benchmark that emphasizes ego-grounding for answers in egocentric videos. Unlike prior VideoQA datasets, the questions in \dataset\ are intentionally diagnostic, probing concepts such as “I”, “my’’ things, “my’’ activities, and “my’’ past interactions. \dataset\ comprises 541 long videos (9.2 minutes on average) and 5K manually annotated %personalized 
questions. Each question is tied to a specific timestamp with also the corresponding answer moment, enabling controlled evaluation in realistic streaming QA settings.

We benchmark a wide range of open-source and proprietary MLLMs. We find that all models struggle: the state-of-the-art GPT-5 achieves only 46.1\% accuracy, which is just over half of human performance at 84.7\% accuracy. Beyond low accuracy, we also observe systematic failure modes: models confuse the camera-wearer with nearby people, mix up personal \vs non-personal objects, answer using only the most salient visible evidence, and frequently ignore temporally distant and visually inconspicuous but necessary context. 
Surprisingly, neither model scaling nor chain-of-thought reasoning yields stable and consistent improvement. To further investigate these failures, we analyze model performances over time and conduct oracle experiments in which models are directly shown key evidence frames. Results show that model performances decay rapidly over time, and providing key frames helps.

For instance, every MLLM tested correctly identifies ``my'' rag on the steering wheel when the question is posed at 26s in Fig.~\ref{fig:intro}. However, all models answer incorrectly when the same question is asked later, after the rag leaves the scene and another person appears with a similar rag. These failures suggest that current models struggle to maintain stable identity- and object-level representations over time. Instead of grounding the reference to my rag, they likely fall back on short-term appearance cues, leading to errors once the true referent is no longer visible. These limitations are further exacerbated by the fact that most models process only 8–32 frames at a time, restricting long-range temporal integration.
Together, our findings show that current MLLMs lack robust ego-grounding and highlight the need for future improvements in long-term memory, temporal tracking, and precise retrieval to support personalized QA in egocentric video. 

Our contributions are threefold:
\begin{itemize}
	\item The first systematic analysis of personalized question answering requiring ego-grounding in MLLMs, evaluating their ability to understand, remember, and reason about the camera-wearer in egocentric videos.
	\item \dataset, a large-scale, diagnostic dataset and benchmark for egocentric personalized VideoQA.
	\item Detailed analyses and insights that reveal limitations of current MLLMs and guidance towards essential research directions for personalized egocentric AI assistants.
\end{itemize}

\section{Related Works}

\textbf{Egocentric VQA.}
Earlier egocentric VQA focused on action recognition \cite{fan2019egovqa} or high-level task understanding \cite{jia2022egotaskqa}, yet with rarely a focus on multiple people interactions or coordination. % without other people present in the scene.
With the push toward egocentric embodied assistance \cite{plizzari2024outlook,wang2023holoassist, plizzari2025spatial, goletto2024amego} and the release of large-scale egocentric datasets (\eg, Ego4D \cite{grauman2022ego4d}), egocentric VQA has attracted growing interest \cite{lin2022egocentric,pramanick2023egovlpv2,mangalam2023egoschema,zhao2023learning,cheng2024egothink,di2024grounded}. However, most of this work focuses on general-purpose visual understanding \cite{mangalam2023egoschema,cheng2024egothink} that mirrors third-person VideoQA settings.

Recent efforts~\cite{yang2025egolife,zhou2025egotextvqa,xiao2025egoblind,yan2025teleego} study assistive egocentric QA aligned with user demands. Yet none specifically target the core challenge of personalized reasoning across extended spatial and temporal contexts. 
Our work fills this gap by providing the first diagnostic study of personalized QA assistance requiring ego-grounding in MLLMs, by focusing on ``me'' (the camera wearer) and ``my'' interacted objects among multiple people and similar distractors.

\textbf{Multimodal LLMs.}
Existing MLLMs primarily address short or third-person video understanding \cite{maaz2023video,zhang2023video,lin2024video,li2024llava}. More recent efforts \cite{lin2024video,zhang2024long,shenlongvu,bai2025qwen2,ye2024mm, di2025streaming,di2024grounded,qian2024streaming} incorporate long-form and egocentric videos for first-person QA, \eg, via extended temporal context windows, hierarchical modeling, memory and retrieval-based modeling. Despite these advances, prior work emphasizes general egocentric comprehension \cite{mangalam2023egoschema,cheng2024egothink} rather than personalized referencing. Ego-grounding requires resolving first-person pronouns and user-specific object references (``I'', ``my rag'', ``my colleague'') over time.  Such capabilities are largely unexamined in current MLLMs and our work is the first systematic evaluation of these personalized reasoning abilities.

\textbf{Ego-cues and Personalized Understanding.}
Egocentric research has extensively explored first-person (ego) cues such as gaze prediction
\cite{lai2024eye,li2013learning,li2025causal}, hand-object interactions
\cite{fan2024benchmarks, xu2023egopca}, and action intent 
\cite{sun2025visual}. These cues provide rich information, but prior works treat them as signals for task or activity understanding
\cite{jia2022egotaskqa, qi2021self}, not as persistent identity anchors. In addition, existing personalized vision-language systems \cite{cohen2022my,yeh2023meta,alaluf2024myvlm,shi2025pvchat} adapt to specific users through additional visual exemplars or textual profiles. In contrast, we aim to reinforcing MLLMs to study personalization directly from past egocentric video itself. 
This requires a form of visual memory and identity tracking under-explored in prior personalization research.

\section{\dataset~Dataset}
\subsection{Task Definition}
\begin{figure*}[t!]
\centering
\includegraphics[width=1.0\textwidth]{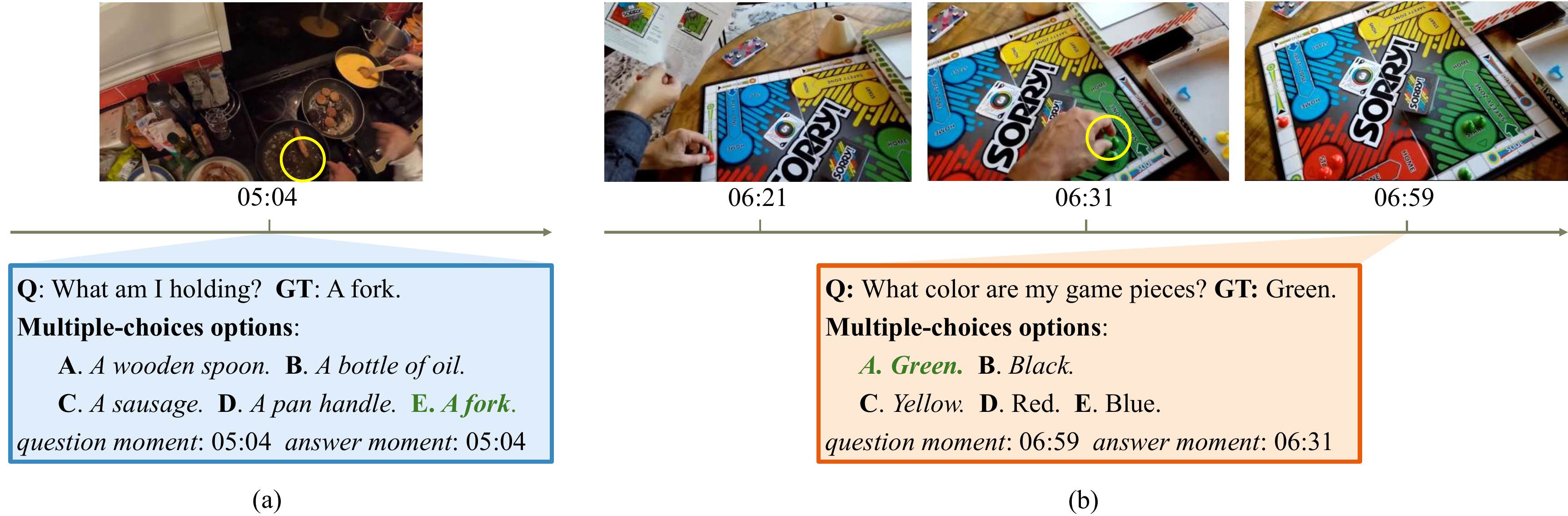}
\vspace{-5mm}
\caption{MyEgo examples. To answer \textit{my} questions, the models must understand (a) which hand is mine, (b) memorize and track the green chess pieces I used (to be distinguished with the red ones of another person, and the blue ones which are not used). }
\label{fig:data_exp}
\vspace{-3mm}
\end{figure*}

We formally define egocentric personalized VideoQA as the task of answering questions that require grounding first-person references in streaming egocentric video. Such questions fall into two categories. The first involves disambiguating ``my'' actions, attributes, or belongings from those of other people in the scene (see \cref{fig:intro}, and \cref{fig:data_exp}(a)). The second involves identifying the specific object the camera wearer interacted with among multiple visually similar instances of the same category (challenging distractors), such as the chess example in \cref{fig:data_exp}(b).

\subsection{Dataset Construction}

\textbf{Video Collection:} 
Our videos are sourced from three public egocentric video datasets: Ego4D~\cite{grauman2022ego4d}, EgoLife~\cite{yang2025egolife}, and CASTEL2024~\cite{rossetto2025castle}. The footage of EgoLife and CASTEL2024 already involve multiple people so we additionally remove single-person videos from Ego4D. EgoLife distributes their videos in 30-second short clips for easy downloading. Also, each frame contains a dynamic timestamp watermark at the upper-left corner. Thus, we construct long videos by concatenating sequential clips from the same recording into approximately 10-minute segments. For watermark, we apply a black mask (blended into the background) to remove it in each frame.
The raw CASTEL2024 videos include long stretches of non-informative content, so we trim them into continuous, activity-focused clips ranging from 6 to 20 minutes. 
After processing, we obtain 541 valid videos of averaging 9.2 minutes, with 182 videos from Ego4D, 257 from EgoLife, and 102 from CASTEL2024.

\textbf{QA Annotation:}
Our preliminary trial shows that generating high-quality, personalized QA pairs for our benchmark was non-trivial. Automatic curation using MLLMs such as GPT-5/Gemini-2.5 Pro were insufficient, as the models often failed to capture the context-specific and personalized nature of our queries. We therefore manually annotated all QA pairs to ensure they accurately reflect the distinctions between actions and objects associated with the camera wearer versus other people in the videos.

We recruited and trained 10 university students to manually annotate the QA pairs based on the video content. 
Annotators were requested to adhere to the following key principles to ensure meaningful and challenging questions: 
\begin{itemize}
\item \textbf{Egocentric}: Question must be framed from the camera wearer's first-person perspective to simulate a direct, personal inquiry. 
\item \textbf{Personalized}: The content must be personalized to highlight distinctions between the camera wearer's actions or objects and those of others, compelling the model to engage in personalized reasoning to first determine, which objects or actions are associated with the camera-wearer, before arriving at a correct answer.
\item \textbf{Visual Answer}: The answers to the questions should be concise and be visible in the videos.
\end{itemize}
The corresponding ground-truth answers are also annotated by the same annotator to ensure their correctness. For each QA pair, we additionally mark the \textit{question moment} when a question is posed and the \textit{answer moment} when the visual evidence for the answer occurs. The \textit{answer moment} is always no later than the corresponding  \textit{question moment}.
After a manual annotation and check, we obtain 5,012 first-person view, personalized QA pairs for open-ended task.

\begin{figure*}[t!]
\centering
\includegraphics[width=1.0\textwidth]{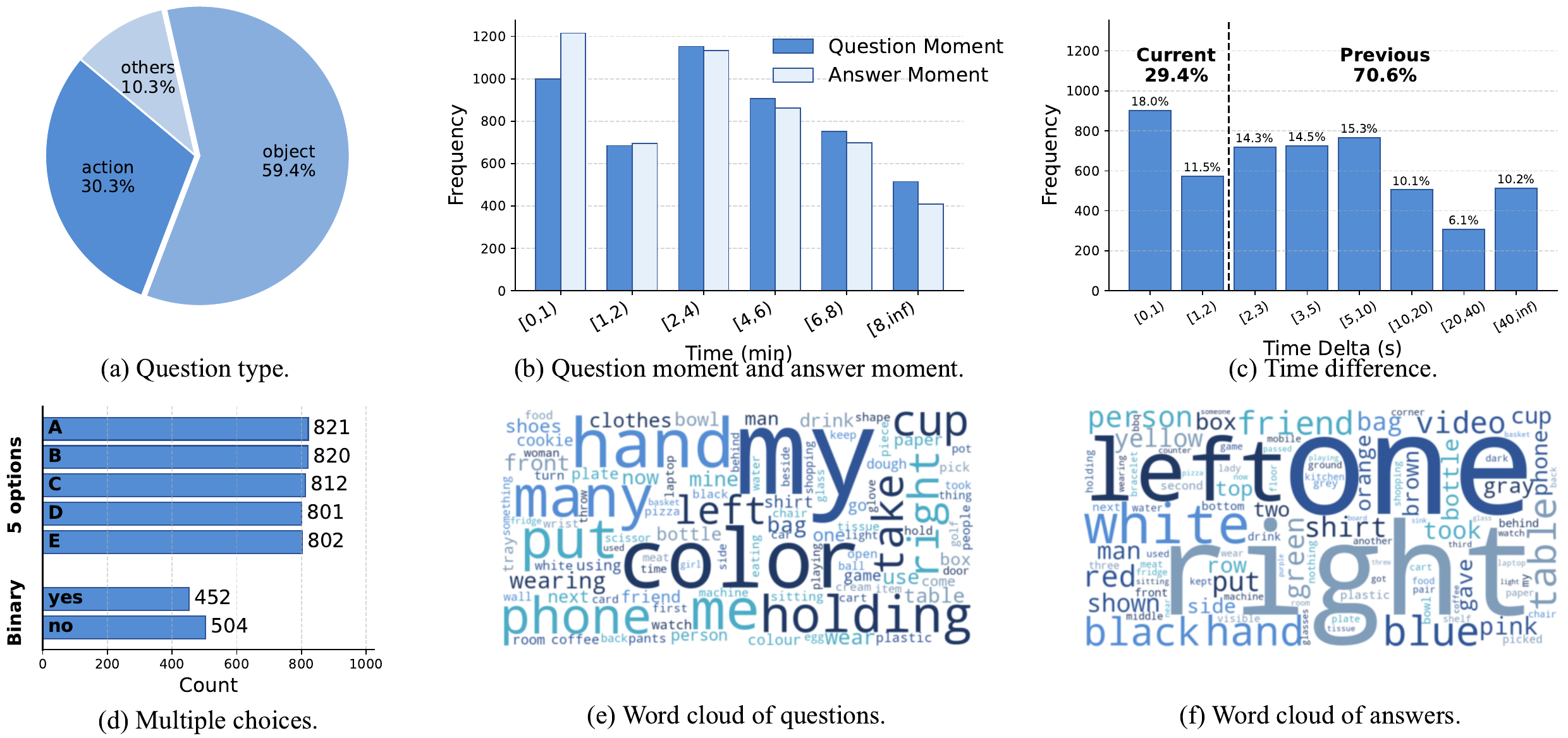}
\caption{Statistic analysis of \dataset}
\label{fig:stat}
\vspace{-3mm}
\end{figure*}

To facilitate more standardized %and objective 
evaluation, we further augment our open-ended QA pairs 
for a multiple-choice (MC) setting.
We feed Gemini-2.5 Pro with the video, question and ground-truth answer, and design a prompt 
to create 4 plausible distractors by requiring that each option describes something verifiably present in the video. Specifically, the prompt prioritized generating distractors that were temporally relevant (appearing at \textit{question moment} or \textit{answer moment}) or contextually confusing (\eg, an action performed by some others versus ``me''). As a fallback, any incorrect event or object appearing in the video is deemed acceptable. For the ``yes/no'' questions, we regard them as 2-option multiple-choice QA task. 

After an initial manual screening for duplicate or invalid options, we add a filtering step to limit data biases. Following \cite{zhang2025mitigatingeasyoptionbias}, we provide only the video frames and answer choices as input, explicitly excluding the question, to both Gemini-2.5 Pro and GPT-5 and identify QA pairs which are solvable by both models. This step filters out simple instances where the correct answer is guessable from the video and options alone. For these rejected QA pairs, we manually refined the distractors based on the videos to ensure a sufficient level of difficulty and specificity on ego-grounding for identifying the correct answers. 

After generation and pre-automatic filtering, we invite 2 students (together with  authors) to carefully inspect and refine through the whole annotations. Our final dataset contains 5,012 questions, each available in both open-ended and multiple-choice tasks, with the latter comprising 953 and 4,059 for 2- and 5-option QA pairs respectively.

\subsection{Dataset Analysis}
\subsubsection{Dataset Statistics}

\cref{fig:stat} presents key statistics of \dataset.  It features 5,012 QA pairs distributed across 541 videos. The videos are on average 9.2 minutes long with an average of 9.3 questions on each. 
The queries in the questions can be categorized semantically as \textbf{Object}, \textbf{Action} and \textbf{Others} (\cref{fig:stat}(a)). Temporally, the question and answer moments are distributed throughout the video (\cref{fig:stat}(b)), with varying temporal separations (\cref{fig:stat}(c)). Questions are considered ``Current'' if the ground truth answer moment is either concurrent or within 2 seconds of the question.  
Questions with answers further back in the history is considered ``Previous''. The average time difference between question and answer moments is about 20 seconds. This average difference already presents a significant challenge, demanding models to recall specific details and  maintain a consistent understanding of the ``me'' concept over time. 
For the multiple-choice (MC) setting, the order of correct choices is randomized to ensure fairness and mitigate positional bias (see \cref{fig:stat}(d)). \cref{fig:stat}(e) and \cref{fig:stat}(f) visualize frequent words in the questions and answers respectively. It shows that the questions strongly feature first-person pronouns (``my'',``me'') and action keywords (``take'',
``put''). Answer keywords are dominated by adjectives describing objective attributes, like colours and locations.

\subsubsection{Dataset Comparison}
\begin{table*}[t!]
    \centering
    \small
    \caption{Benchmark comparison. \dataset~highlights ego-grounding and memorizing the camera wearers from ego video stream where multiple people and multiple similar objects present. MP: questions require distinguishing ego-person from other persons. MO: questions require distinguishing objected associated with ego-person from other similar-looking objects. OE/MC: Open-Ended/Multi-Choice.}
    \label{tab:comparison}
    \setlength{\tabcolsep}{9.3pt}
    \begin{tabular}{l ccccccc}
        \toprule
        \textbf{Benchmark} & \textbf{\#Q} & \textbf{\#V} & \textbf{Ave. Len} & \textbf{Challenges} & \textbf{MP} & \textbf{MO} & \textbf{Task} \\
        \midrule
         QAEgo4D \cite{qaego4d} &1,854& 166 &8.2 min&Episodic Memory&\xmark&\xmark&OE\\
         EgoSchema\cite{mangalam2023egoschema}&5,063&5,063&3 min&Long Ego Video Understanding &\xmark&\xmark&MC\\
         EgoThink\cite{cheng2024egothink}&700&595&-&General Ego Vision Understanding&\xmark&\xmark&OE\\
         EgoMemoria\cite{ye2024mm}&7,026&629&0.5 to 60 min&General Ego Video Understanding&\xmark&\xmark&MC\\
         EgolifeQA \cite{yang2025egolife}&6,000&6&44.3 h&Multimodal Episodic Memory&\xmark&\xmark&MC\\
         EgoTextVQA \cite{zhou2025egotextvqa}&7,064&1,507& 1.7 min&Ego Scene-Text Understanding&\xmark&\xmark&MC\\
         EgoBlind \cite{xiao2025egoblind}&5,311&1,392& 40s & Blind Assistance &\xmark&\xmark&OE\\
        \rowcolor{gray!20} % Highlight our work
        \dataset~(Ours) & 5,012 & 541 & 9.2 min & Personalized Understanding & \cmark  & \cmark & OE/MC   \\
        \bottomrule
        \end{tabular}
        % }
\end{table*}
We compare \dataset~with several egocentric video QA benchmarks in \cref{tab:comparison}. As highlighted, \dataset~features personalized questions that emphasizes ego-grounding and tracking the camera wearers and their interactions
in complicated environments where multiple people and similar-looking objects present. Prior datasets, such as EgoSchema \cite{mangalam2023egoschema}, EgoMemoria \cite{ye2024mm}, and EgoThink \cite{cheng2024egothink}, primarily focus on general-purpose ego-vision understanding.
While QAEgo4D \cite{qaego4d} and EgoLifeQA \cite{yang2025egolife} emphasize episodic memory, they do not require linking the recalled moment to help ego-ground the camera wearer in the current scene, which is underscored in \dataset. 
Specialized datasets like EgoTextVQA \cite{zhou2025egotextvqa} and EgoBlind \cite{xiao2025egoblind} 
have concentrated on specific functionalities like understanding scene text or assisting the visually impaired in egocentric manner. \dataset~shares similar goal in egocentric assistance, but differs in the emphasis on ego-grounding for personalized QA in long video stream.
It specifically tests the model capacity to analyze, understand, and remember the camera wearer's trajectory and intent over time, making it better reflect the demands of personalized embodied QA assistants.

\section{Experiments}
\label{sec:formatting}

\subsection{Experimental Setups}
\textbf{Evaluation.} For open-ended (OE) QA, we prompt GPT-5 mini  \cite{openai2025gpt} as an evaluator to give a binary judgment (``yes'' or ``no'') on whether a model's response matches the ground truth (GT) answer, to obtain the Accuracy (0-100\%, the percentage of ``yes'' answers in evaluation). Meanwhile, we obtain the match Score (0-5, 5 indicates an exact match.) that signals the match extent of two answers. After refining the evaluation prompt for GPT-5 mini, we achieve an agreement rate of 94\% with human judgment (see Supplementary), demonstrating a reasonable automatic evaluation mechanism. 
For more stable evaluation, we also report QA accuracy on multiple-choice (MC) setting. Specifically, we provide candidate answers for each question, and prompt the models to output the selected answer. Detailed prompts for QA and evaluation are given in the Supplementary.

\textbf{Models.} We analyze both popular closed-source and open-source MLLMs. For closed-source ones, we evaluate Gemini-2.5 Pro  \cite{comanici2025gemini} and GPT-5  \cite{openai2025gpt}. For open-source models, we collect them from three groups: \textbf{1) General-purpose understanding:} Qwen2.5-VL series  \cite{bai2025qwen2}, Qwen3-VL series  \cite{Qwen3-VL}, InternVL2.5-8B  \cite{chen2024far}, Intern3-VL series  \cite{zhu2025internvl3}, Intern3.5-VL series  \cite{wang2025internvl3}, LLaVA-OneVision  \cite{li2024llava}, LLaVA-Video  \cite{zhang2024video}, MiniCPM-V 4.5 \cite{yu2025minicpm}, \textbf{2) Long video understanding:} LongVA  \cite{zhang2024longva}, LongVU  \cite{shenlongvu}, and \textbf{3) Streaming video understanding:} Flash-VStream \cite{zhang2025flashvstream} and Dispider \cite{qian2025dispider}. Additionally, to benchmark human performance, we evaluated two university students who were not involved in annotation on a random subset of 300 samples (6\% of the full dataset). Related results are presented in Tab.~\ref{tab:main_results} and Supplementary Tab.~\ref{tab:all_results}.

\textbf{Implementation.} For each question, we uniformly sample a fixed number of video frames up to the question timestamp, with a maximum rate of 1 fps. While other models default to 32 frames, LongVA and LongVU are supplied with 128 frames. Furthermore, we explicitly prompt the model to adopt a first-person perspective, focusing on the camera wearer's (\textit{my}) objects and actions. 

\subsection{Main Results}

\begin{table*}[ht!]
  \centering
  \small
  \setlength{\tabcolsep}{4.5pt}
  \caption{Evaluation results on \dataset\ dataset. We present the performance on both multiple-choice (MC) and open-ended (OE) tasks. MCQA includes binary (MC-2) and five-option (MC-5) QA. For OEQA, we analyze the results across different categories and temporal locality. Cur.: question whose answer can be found in the current moment (within 2s). Pre.: question whose answer is located in the past video content. 480P: resize the height to be 480.}
  \label{tab:main_results}
  \renewcommand{\arraystretch}{0.8}{
 \begin{tabular}{l c| c c|c c| c| ccc| c  c| c }
\toprule
\multirow{2}{*}{\textbf{Methods}} & \multirow{2}{*}{\textbf{Res.}} 
& \multicolumn{5}{c|}{\textbf{Multiple-Choice}} 
& \multicolumn{6}{c}{\textbf{Open-Ended}} \\
\cmidrule(lr){3-7} \cmidrule(lr){8-13}
& & MC-2 & MC-5 & Cur. & Pre.& Avg. 
& Act. & Obj. & Others & Cur. & Pre. & Avg. (Acc/Score) \\
\midrule
\rowcolor{gray!40}
\textbf{Human} & - & 95.1 & 92.1 &93.4&92.3& 92.7 & 82.4 & 84.8 & 91.2 & 84.0 & 85.0 & 84.7 \\
\midrule
\multicolumn{13}{c}{\textit{Closed-source Models}} \\
\midrule
GPT-5 \cite{openai2025gpt} & 480P & \textbf{66.4} & \textbf{53.7} &  \textbf{56.9}&  \textbf{55.8}&\textbf{56.1} & \textbf{50.0} & \textbf{44.6} & \underline{43.1} & \textbf{51.1} & \textbf{44.0}& \textbf{46.1} / \textbf{2.5} \\
Gemini-2.5 Pro \cite{comanici2025gemini} & 720P & \underline{61.8} & \underline{45.5} & \underline{49.0} & \underline{48.4} & \underline{48.6} & \underline{40.2} & \underline{40.2} & \textbf{47.7} & \underline{42.4} & \underline{40.3} & \underline{40.9} / \underline{2.2} \\
\midrule
\multicolumn{13}{c}{\textit{Open-source Models}} \\
\midrule
InternVL2.5-8B \cite{chen2024far} & $448^2$ & 53.1 & 36.6&41.3&39.1 & 39.8 & 27.8 & 23.1 & 24.1 & 27.2 & 23.5 & 24.5 / 1.4 \\
InternVL3.5-8B-Instruct \cite{wang2025internvl3} & $448^2$ & 54.6 & 36.6 &40.5&39.8& 40.0 & 31.9 & 28.6 & 31.9 & 30.3 & 29.7 & 29.9 / 1.6 \\
LongVA \cite{zhang2024longva} & $336^2$ & 56.9 & 30.9&40.2&34.0 & 35.8 & 33.4 & 31.3 & 32.4 & 34.6 & 31.0 & 32.1 / 1.7 \\
InternVL3.5-8B-Thinking \cite{wang2025internvl3} & $448^2$ & 54.5 & 37.7&41.8&40.5 & 40.9 & 32.5 & 32.2 & 33.4 & 32.5 & 32.4 & 32.4 / 1.8 \\
LongVU \cite{shenlongvu} & ori. & 53.3 & 32.8&38.0&36.2& 36.7 & 33.3 & 31.7 & 34.6 & 33.2 & 32.2 & 32.5 / 1.8 \\
LLaVA-OneVision \cite{li2024llava} & $384^2$ & 53.5 & 34.5 &40.9&36.9& 38.1 & 34.5 & 31.8 & 40.5 & 35.7 & 32.6 & 33.5 / 1.9 \\
MiniCPM-V 4.5 \cite{yu2025minicpm} & ori. & 53.8 & 36.1 &41.4&38.7& 39.5 & 36.2 & 32.1 & 39.0 & 35.3 & 33.5 & 34.0 / 1.9 \\
InternVL3-8B \cite{zhu2025internvl3} & $448^2$ & 54.5 & 38.4&42.4&41.0 & 41.4 & 34.7 & 33.3 & 38.6 & 34.7 & 34.1 & 34.3 / 1.8 \\
Qwen2.5-VL-7B \cite{bai2025qwen2} & ori. &55.1 & 32.3&40.8&34.9& 36.6 & 35.4 & 33.7 & 34.8 & 37.4 & 33.0 & 34.3 / 1.8 \\
Qwen3-VL-8B-Thinking \cite{Qwen3-VL} & ori. & 56.2 & 31.5 &38.9&35.1& 36.2 & 36.4 & 33.1 & 35.1 & 36.6 & 33.4 & 34.3 / 1.8 \\
LLaVA-Video \cite{zhang2024video} & ori. & 54.8 & 36.0 &43.0&38.2& 39.6 & 35.2 & 34.0 & 40.0 & 37.4 & 33.9 & 35.0 / 1.9 \\
Qwen3-VL-8B-Instruct \cite{Qwen3-VL} & ori. & 55.0 & 36.6&41.4&39.5& 40.1 & 38.5 & 35.5 & 35.7 & 37.4 & 36.0 & 36.4 / 2.0 \\
\bottomrule
\end{tabular}
}
\end{table*}

\begin{figure*}[!t]
\centering
\includegraphics[width=1.0\textwidth]{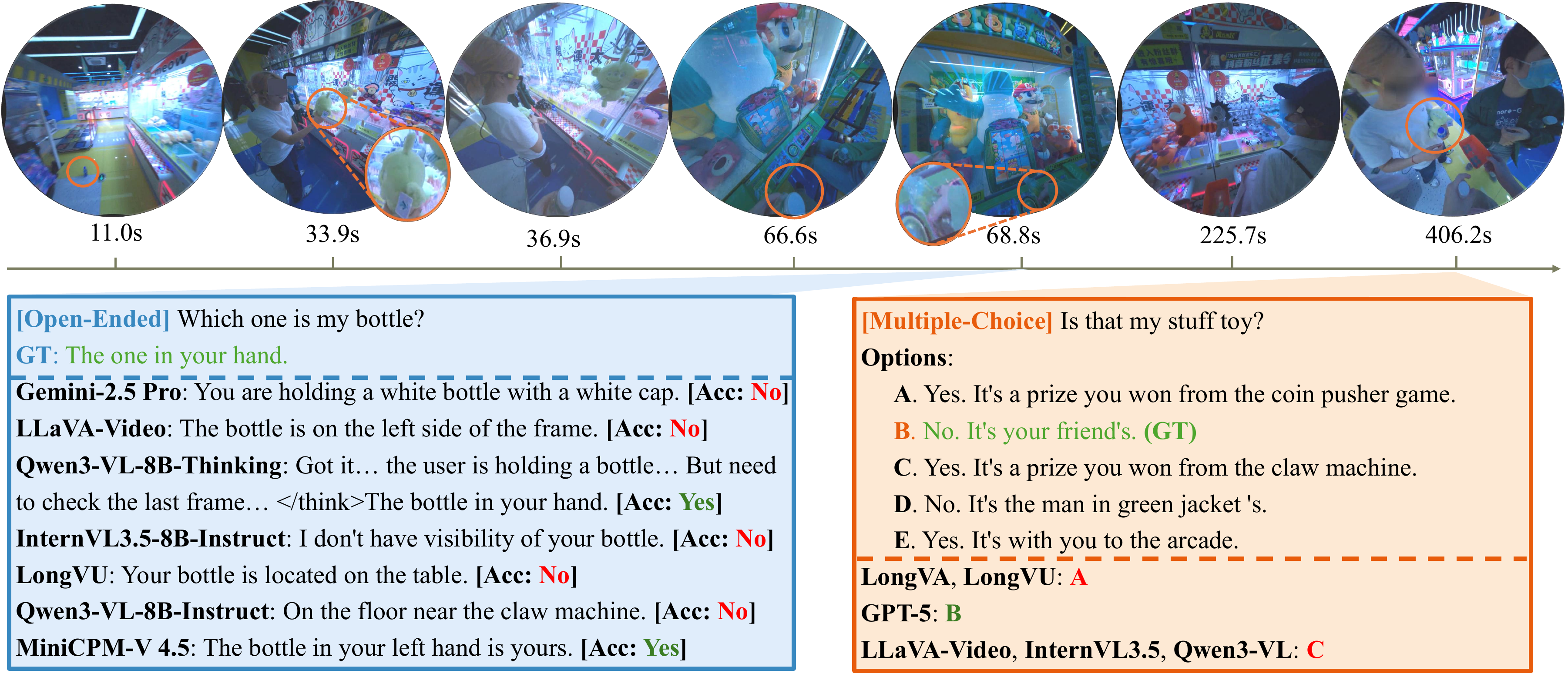}
\caption{Result visualization.}
\label{fig:reexp}
\vspace{-3mm}
\end{figure*}

\cref{tab:main_results} and \cref{fig:reexp} present the performances of diverse models on \dataset. We summarize the following points. % \, benchmark. 

\textbf{General Observations.} Both open- and closed-source models lag behind human performance by 33\%$\sim$55\%, underscoring the challenging nature of \dataset. GPT-5 generally outperforms others in terms of overall accuracy, though no single model consistently excels across all categories. Among the open-source models, InternVL3 wins in MCQA while Qwen3-VL champions in OEQA. Noteworthy, all models, especially for the InternVL series of models, exhibit substantial performance drops when transferred from the multiple-choice setting to open-ended QA, suggesting that much of their multiple-choice success may rely on answer-choice shortcuts rather than engaging in faithful multimodal reasoning. However, 
MC questions, although simpler than OE ones, are still quite challenging and most models only achieve chance-level (50\%) performances on binary (MC-2) questions. This is because our distractor answers are specially curated to be misleading if without truly grounding the questioner (camera wearer) in the video. 

Additionally,  In the Supplementary, we analyze MLLMs of different parameter sizes and find that larger models do not consistently outperform the smaller ones. Even within the same model family, smaller variants (\eg~4B) can achieve competitive or even superior performance. This observation indicates that general-purpose model scaling cannot solve the challenge in \dataset.

\textbf{Failure Instances.} Unsurprisingly, questions whose answers are not found in the question moments are more challenging (``Previous'' vs. ``Current'' category), suggesting that the models struggle with grounding and tracking the camera wearers and their associated things. For instance, Gemini-2.5 Pro~\cite{comanici2025gemini} and Qwen3-VL-8B-Instruct~\cite{Qwen3-VL} merely describe question-irrelevant objects presented in current or past moment (see \cref{fig:reexp}, left), while 
LLaVA-Video~\cite{zhang2024video} and InternVL3.5~\cite{wang2025internvl3} series tend to summarize the global events in specific frames (see \cref{fig:reexp}, right) 
without linking them to the camera wearer’s trajectory, leading to incorrect answers.

\textbf{Counter-intuitive Findings.} Unexpectedly, 
``thinking'' models such as Qwen3-VL-8B-Thinking~\cite{Qwen3-VL} and InternVL3.5-8B-Thinking offer no significant accuracy gains. 
This result conflicts with the observations in existing research \cite{li2025videochat}, which shows that MLLMs benefit from long-CoT reasoning on general video understanding task, highlighting the unique challenge of ego-grounding for personalized understanding in \dataset.
Models using more input frames (e.g., LongVA~\cite{zhang2024longva} and LongVU~\cite{shenlongvu} with 128 vs. the default 32) also show no improvement on either Current or Previous questions. We speculate that modeling more frames will introduce increased noises that in turn will contaminate the often only partially visible ego cues that are key to identify the camera wearers. Finally, scaling up model parameters only provides a marginal yet unstable accuracy increase (see Supplementary). 

\subsection{Controlled Analyses}
\label{sec:diag}

\begin{table}[!t]
    \small
    \centering
  \caption{Effects of \textit{answer moment} as input. Uni.: uniform sampling. Q\&A: key frames at QA moments as input.}
  \label{tab:ques_frame}
  \setlength{\tabcolsep}{4.6pt}
  \renewcommand{\arraystretch}{0.8}{
  \begin{tabular}{lcll}
    \toprule
    \textbf{Models} &\textbf{Input}&\textbf{Acc@Cur.}&\textbf{Acc@Pre.}\\
    \toprule
    Qwen3-VL-8B-Think \cite{Qwen3-VL}&\makecell[c]{Uni.\\Q\&A}&\makecell[c]{38.4\\41.3 {\footnotesize \textcolor{ForestGreen}{ $\uparrow$ 2.9}}}&\makecell[c]{32.0\\41.1{\footnotesize \textcolor{ForestGreen}{ $\uparrow$ 9.1}}}\\
        \midrule
    {Qwen3-VL-8B-Instruct} \cite{Qwen3-VL}&\makecell[c]{Uni.\\Q\&A}&\makecell[c]{36.9\\46.3 {\footnotesize  \textcolor{ForestGreen}{ $\uparrow$ 9.4}}}&\makecell[c]{35.5\\41.4{\footnotesize \textcolor{ForestGreen}{ $\uparrow$ 5.9}}}\\
    \midrule
    InternVL3-8B \cite{zhu2025internvl3}&\makecell[c]{Uni.\\Q\&A}&\makecell[c]{35.6\\41.3 {\footnotesize \textcolor{ForestGreen}{ $\uparrow$ 5.7}}}&\makecell[c]{34.8\\40.4 {\footnotesize \textcolor{ForestGreen}{ $\uparrow$ 5.6}}}\\
    \midrule
     LLaVA-Video \cite{zhang2024video}&\makecell[c]{Uni.\\Q\&A}&\makecell[c]{36.3\\37.7 {\footnotesize \textcolor{ForestGreen}{ $\uparrow$ 1.4}}}&\makecell[c]{33.7\\41.6 {\footnotesize \textcolor{ForestGreen}{ $\uparrow$ 7.9}}}\\
        \midrule
    Qwen2.5-VL-7B \cite{bai2025qwen2}&\makecell[c]{Uni.\\Q\&A}&\makecell[c]{37.7\\43.4 {\footnotesize \textcolor{ForestGreen}{ $\uparrow$ 5.7}}}&\makecell[c]{33.2\\42.4 {\footnotesize \textcolor{ForestGreen}{ $\uparrow$ 9.2}}}\\
        \midrule
    Gemini-2.5 Pro \cite{comanici2025gemini}&\makecell[c]{Uni.\\Q\&A}&\makecell[c]{42.4\\49.3 {\footnotesize \textcolor{ForestGreen}{ $\uparrow$ 6.9}}}&\makecell[c]{40.3\\51.5 {\footnotesize \textcolor{ForestGreen}{ $\uparrow$ 11.2}}}\\
    
    \bottomrule
  \end{tabular}
  }
  \vspace{-3mm}
\end{table}

\begin{figure}
\centering
\includegraphics[width=0.48\textwidth]{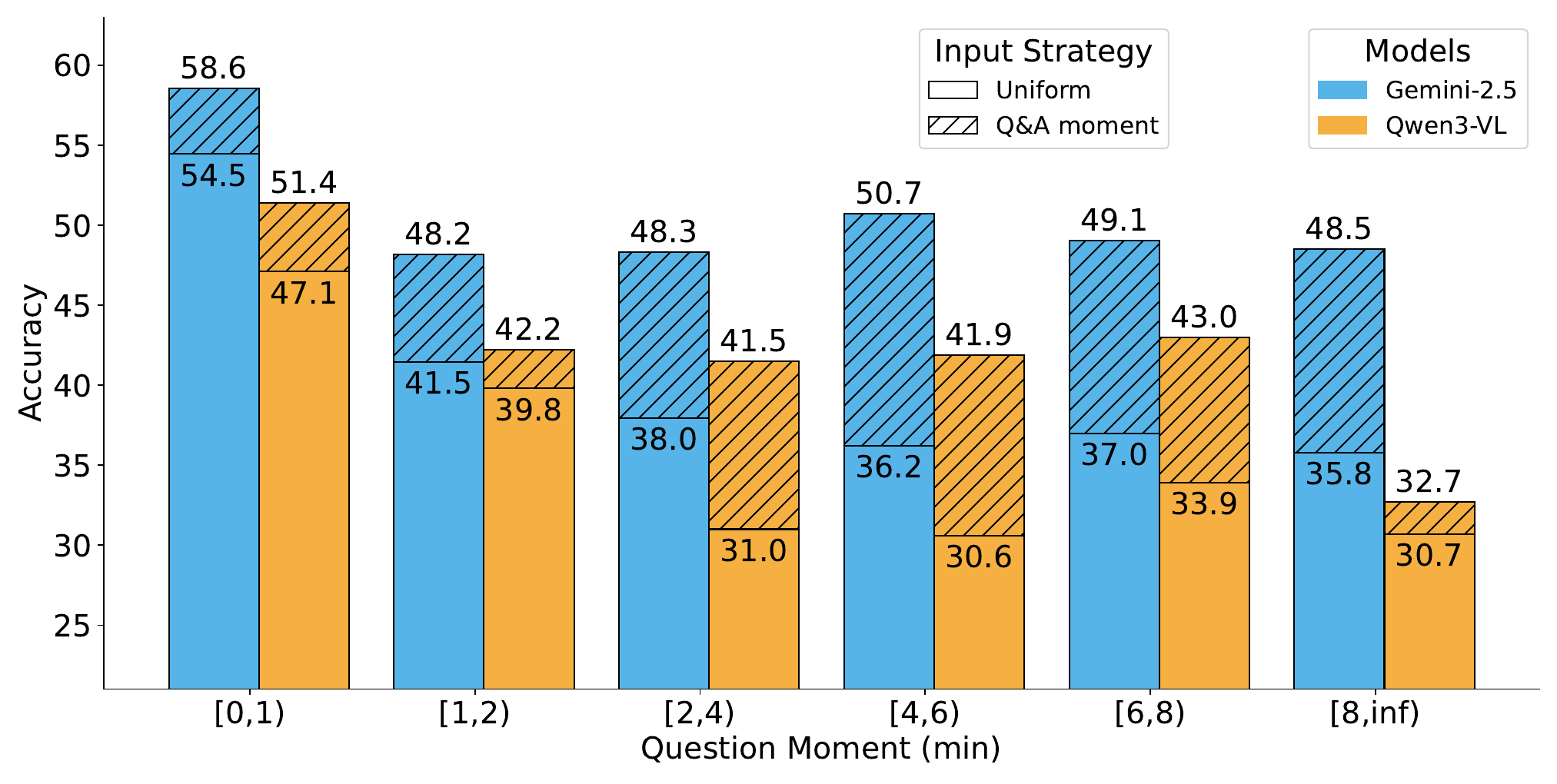}
\caption{Accuracy distribution along the video stream. Accuracy drops along time with uniform sampling but remains relatively stable with GT moment input.}
\label{fig:distri}
\vspace{-3mm}
\end{figure}
\noindent\textbf{Memorizing key information is essential for our task.} We investigate a \textit{question moment} and \textit{answer moment} aware sampling strategy (Q\&A strategy), where we uniformly sample 8 frames from the \textpm 1.5s intervals surrounding both the \textit{answer moment} and \textit{question moment}, and concatenate them to form the final 16-frame input. If the two intervals overlap, we combine them into a single interval from which 16 frames are uniformly sampled. \cref{tab:ques_frame} show that taking such a strategy significantly improves the models' performance, though with fewer frames input. The improvement is more dramatic for ``Previous'' questions, where Gemini-2.5 pro \cite{comanici2025gemini} and Qwen2.5-VL-7B \cite{bai2025qwen2} gain 11.2\% and 9.2\% respectively. It is likely because correctly answering ``Previous'' questions requires accurately locating and distinguishing key information %(especially those highly related to \textit{me}) 
from tens of seconds or even minutes ago, while uniform sampling may miss this information or introducing too much irrelevant background information. Interestingly, accuracy on ``Current'' questions also improves by 1.4\% to {9.4\%}, which uniform sampling should already provide sufficient information to answer. This may be because the Q\&A input provides a larger number of key frames compared to uniform sampling (16 versus \textasciitilde 1). Furthermore, information from irrelevant and redundant frames in uniform sampling can interfere with the model's judgment.

We also analyze the distribution of accuracy along the video stream under both uniform sampling and GT-moment sampling strategies in \cref{fig:distri}. Intuitively, with a fixed number of uniformly sampled frames, models often trade off key details for long-ranged modeling, thus leading to lower accuracy as ego cues are often partially visible. Our experimental results are consistent with this intuition: model accuracy is the highest for questions asked within the first minute and declines as the video stream progressed, with the lowest performance for questions posed after 8 minutes, which highlights the inherent challenge of long-ranged video QA in \dataset. Crucially, the Q\&A strategy consistently outperforms the uniform sampling method across every time bin for both models, once again demonstrating the importance of key frames grounding. Overall, the improvements suggest that enabling the model to distinguish key, self-related information, filter out irrelevant visual clutter, 
and focus on memorizing egocentric details, especially in long-form video QA scenarios, is promising for future exploration.\\

\begin{figure}
\centering
\includegraphics[width=0.48\textwidth]{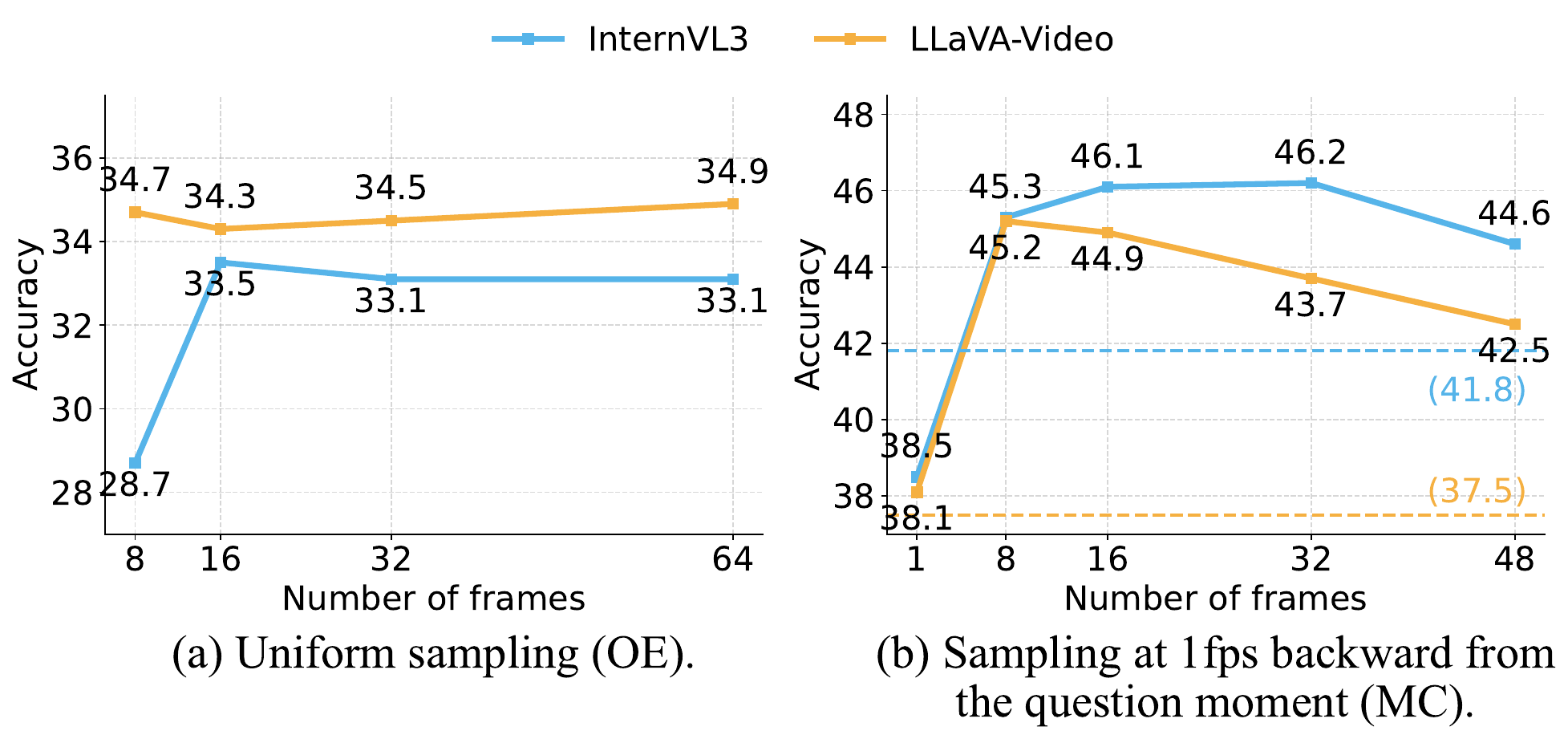}
\caption{Effects with different number of frame inputs. The dashed lines in (b) are the results of uniformly sampling 32 frames.}
\label{fig:frame}
\vspace{-3mm}
\end{figure}

\textbf{More frames do not always bring better performance.}
We feed InternVL3-8B \cite{zhu2025internvl3} and LLaVA-Video \cite{zhang2024video} with a varying number of frames, ranging from 8 to 64, as shown in \cref{fig:frame}(a). For open-ended QA, the performance of InternVL3-8B peaks with 16 frames and then slightly declines, while only reaches 28.7\% with 8 frames. LLaVA-Video's accuracy remains relatively stable, staying within the range of 34-35\%. These suggest that adding more frames alone under uniform sampling strategy does not boost performance. Furthermore, inspired by the nature of human asking questions shortly after an event, and by the observation that in 90\% of our QA pairs the time difference between \textit{question moment} and \textit{answer moment} is less than 40 seconds (see \cref{fig:stat}), we explored a backward sampling strategy. Specifically, for multiple-choice setting, we sample 1 to 48 frames backward from the \textit{question moment} at 1 fps. \cref{fig:frame}(b) shows that such a strategy surpasses the baseline accuracy achieved by uniformly sampling (indicated by the dashed lines), though with much fewer frames. However, the benefits are not linear. The most significant improvement occurs when increasing the input frames from 1 to 8, yielding accuracy improvement of 6.8\% and 7.1\% on InternVL3 and LLaVA-Video, respectively. Then the accuracy enters a plateau, with only minor fluctuations or even a slight decrease, possibly due to redundant frames that do not necessarily for better understanding. These findings underscore that the relevance of visual information is more critical than quantity, inspiring future study into more intelligent moment detection.

\begin{table}[!t]
    \small
    \centering
  \caption{Effects of different prompt modification. We test several models' performance with a prompt where cues of "personalization" is removed ("Remove") and questions where first-person pronouns are replaced with "the camera wearer" ("Enhanced"), respectively. We select the thinking version of InternVL3.5-8B here.}
  \label{tab:prompt}
  \begin{tabular}{lccc}
    \toprule
    \textbf{Models} &\textbf{Input}&\textbf{OE}&\textbf{MC}\\
    \toprule
    InternVL3.5-8B \cite{wang2025internvl3}&\makecell[c]{Original\\Enhanced\\Remove}&\makecell[c]{33.1\\33.2 \textcolor{ForestGreen}{ $\uparrow$  0.1}\\ 31.7 \textcolor{Red}{ $\downarrow$ 1.4}}&\makecell[c]{39.5\\41.1 \textcolor{ForestGreen}{ $\uparrow$ 1.6}\\ 38.5 \textcolor{Red}{ $\downarrow$ 1.0}}\\
    \midrule
    InternVL3-8B \cite{zhu2025internvl3}&\makecell[c]{Original\\Enhanced\\Remove}&\makecell[c]{35.0\\34.7 \textcolor{Red}{ $\downarrow$ 0.3}\\ 33.7 \textcolor{Red}{ $\downarrow$ 1.3}}&\makecell[c]{41.8\\40.4 \textcolor{Red}{ $\downarrow$ 1.4}\\ 41.4 \textcolor{Red}{ $\downarrow$ 0.4}}\\
        \midrule
    LLaVA-Video \cite{zhang2024video}&\makecell[c]{Original\\Enhanced\\Remove}&\makecell[c]{34.5\\35.9 \textcolor{ForestGreen}{ $\uparrow$ 1.4}\\ 35.3 \textcolor{ForestGreen}{ $\uparrow$ 0.8}}&\makecell[c]{37.5\\38.1 \textcolor{ForestGreen}{ $\uparrow$ 0.6}\\ 37.5 \textcolor{ForestGreen}{ $\uparrow$ 0.0}}\\
    \bottomrule
  \end{tabular}
\end{table}

\begin{figure}[t!]
\centering
\includegraphics[width=0.5\textwidth]{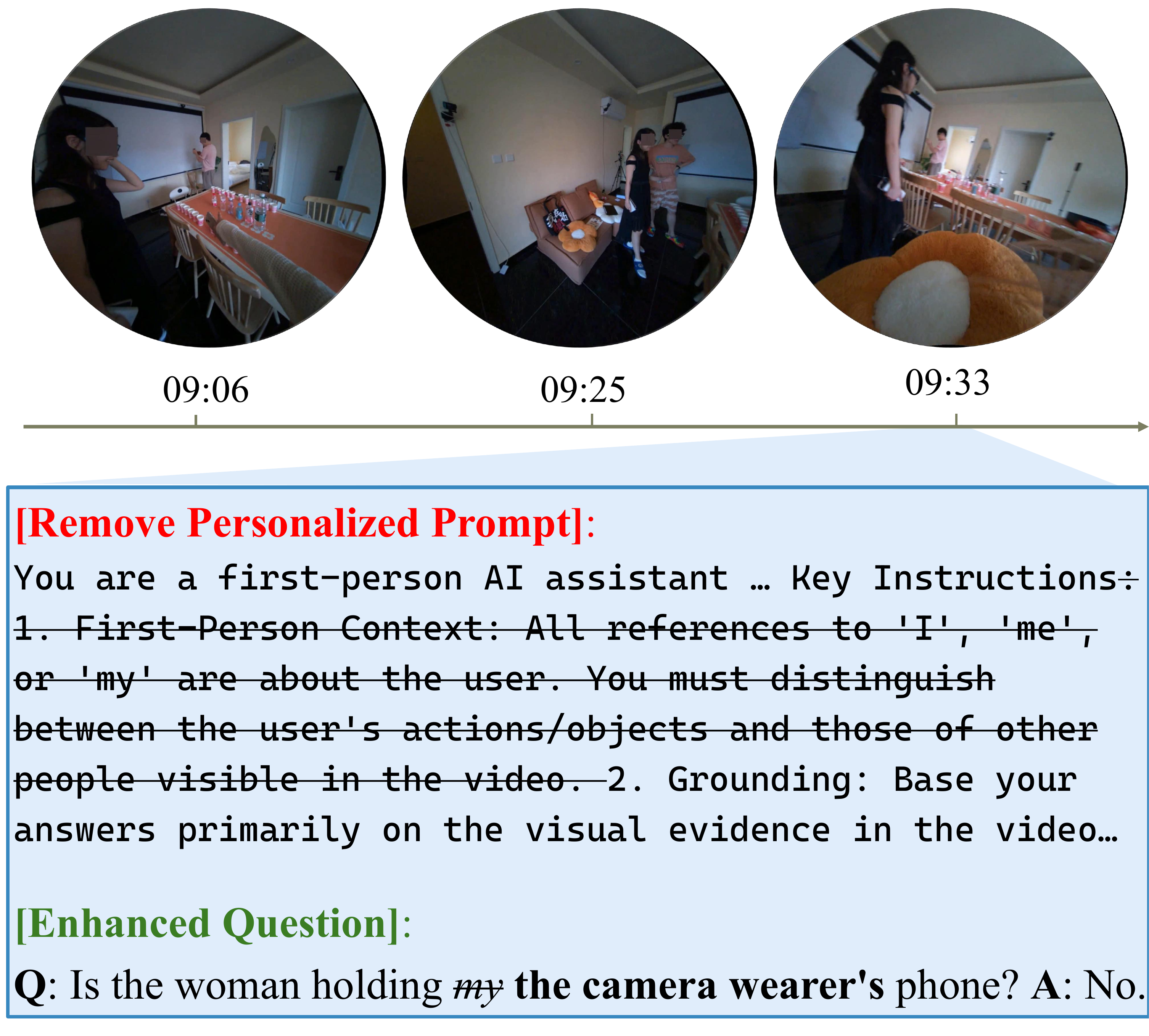}
\caption{Example prompts of (top) removing the personalized cues and (bottom) enhancing the clarification by reminding that 'I' refers to the camera wearer.}
\label{fig:prompt}
\vspace{-3mm}
\end{figure}

\textbf{Personalization-aware prompting has a small but measurable impact.}  To evaluate the impact of different prompts on model performance, we introduce two systematic modifications to the personalized prompt. Specifically, we test an ``Enhanced'' prompt, where first-person pronouns (e.g., ``I'', ``my'') in the questions are replaced with ``the camera wearer ('s)'', and a ``Remove'' prompt, which removes personalization cues entirely (see \cref{fig:prompt}). Results are concluded in \cref{tab:prompt}. 
The ``Enhanced'' strategy generally improves performance, but not always.  
For instance, while LLaVA-Video \cite{zhang2024video} sees a consistent improvement of over 1\% in both OE and MC tasks, the performance of InternVL3.5-8B \cite{wang2025internvl3} remains largely unchanged. Conversely, removing personalization predominantly degrades performance, as seen in the notable 1.4-1.6\% accuracy drop of the InternVL3.5-8B model. However, its impact can be nuanced, occasionally resulting in slight performance gains for some models. These findings suggest that the models are not drastically sensitive to these specific prompt alterations, but a clear pattern indicates that enhancing the prompt (\textit{i.e.,} reminding that ``I'' refers to the camera wearer in the question) is generally helpful. 

\section{Conclusion}
We studied whether MLLMs can faithfully understand ``me'', remember ``my'' past interactions, and track ``my'' context over extended spatial and temporal scales. We introduced \dataset, the first egocentric VideoQA dataset designed for personalized user questions requiring ego-grounding—distinguishing me and my objects from others. Using \dataset, we conducted a comprehensive evaluation of modern MLLMs and found that all models struggle, and even scaling and “thinking” offer only limited gains. Models perform better when key moments are provided or lie close to the question time, yet large gaps remain compared with human performance, and their accuracy drops sharply once the answer leaves the current view. This indicates that while models can partially understand me, they fail to truly remember me. These findings highlight the need for stronger long-range memory in the short term and genuinely personalized reasoning in the long term. We hope \dataset\ and related analyses help drive progress toward these goals.

\section*{Acknowledgements}
This research is supported by the Ministry of Education, Singapore, under its MOE Academic Research Fund Tier 2 (MOE-T2EP20125-0037).

{
    \small
    \bibliographystyle{ieeenat_fullname}
    \bibliography{main}
}
\clearpage
\appendix
\maketitlesupplementary

\section{\dataset~Dataset}
In this part, we will introduce additional details about the construction of \dataset~dataset, including the video source datasets and the multiple-choice generation pipeline. 
Our dataset and related resources are available at \href{https://github.com/Ryougetsu3606/MyEgo}{https://github.com/Ryougetsu3606/MyEgo}.
\subsection{Video Source Introduction}
\textbf{Ego4D}~\cite{grauman2022ego4d}~offers 3,670 hours of daily activity video spanning diverse scenarios: household, outdoor, workplace, and etc, which is captured by 931 unique camera wearers from 74 worldwide locations. We select 182 videos from QAEgo4D \cite{qaego4d} subset, with each video involving two or more people engaging in the same scene or event.

\noindent\textbf{EgoLife}~\cite{yang2025egolife}~consists of 44-hour egocentric videos recording a week of shared living experience of six young volunteers. To better raise questions related to multi-person scenarios, we further select clips containing rich group activities, including videos from all six volunteers on the same day (Day 3), as well as footage from a single volunteer (A4) spanning four days (Day 3 to Day 6). We then concatenate the sequential short clips into 257 longer ones, with each about 10 minutes.

\noindent\textbf{Castle2024}~\cite{rossetto2025castle}~is a large-scale, multimodal dataset designed for advancing study in lifelogging, human activity analysis and multimodal retrieval. The entire set contains over 600 hours of videos from 15 time-aligned recorders, including 10 participants and 5 fixed cameras. We select a ~30-hour subset of videos and segment them into clips of 6–20 minutes. We additionally filter out clips that lack multiple people or contain incomplete recording data, and eventually obtain 102 valid videos.

\subsection{Multiple Choice Generation}
\cref{tab:distractor_gen} presents the prompt used for distractor generation. After obtaining the generated distractors, we first refine the options to produce concise expressions consistent with the style of the correct answers, preventing answer leakage due to formatting discrepancies. To further limit option bias, we manually revise QAs that both Gemini-2.5 Pro~\cite{comanici2025gemini} and GPT-5~\cite{openai2025gpt} can correctly answer even without access to the video or question. For human participation, we particularly ensure the negative options be misleading unless the models truly understand the video contents available at the question moments and correctly ground ``me'', ``my things'', and ``my past'' in the egocentric scene.

\subsection{Question Categories}
For better evaluation and analysis, we categorize the questions into 3 groups: 1) Action: questions focus on distinguishing my actions from those of other people nearby, 2) Object: questions focus on distinguishing my objects from other similar objects in the scene, and 3) Others: other questions that are not covered by the previous two categories. Examples of questions for each category are presented in \cref{tab:qtype}. Additionally, we split the questions according to whether their answers are visible at the question moments: 1) Current: answer is visible at the question moment, and 2) Previous: answer is not visible at the question moment and demands retrieving past video contents.

\begin{table}[t!]
    \centering
    \small
    \caption{Question categories and examples of \dataset.}
    \begin{adjustbox}{max width=\linewidth}
    \begin{tabular}{c|c|l}
    \toprule
    \textbf{Category} & \textbf{Number/Ratio}& \makecell[c]{\textbf{Question Examples}}\\
    \hline
    \textbf{Action}  & 1,519/30.3\%  & \makecell[l]{What am I \emph{holding}? \\ Did I \emph{give} her my power bank? \\ Where did I \emph{put} my pan? } \\
    \hline
    \textbf{Object} & 2,975/59.4\% & \makecell[l]{Is this my rag? \\Where is my screwdriver? \\What color is my helmet?} \\
    \hline
    \textbf{Others} & 518/10.3\% & \makecell[l]{Is this bowl the same as before? \\ Did I get more than 200 scores?} \\
    \bottomrule
    \end{tabular}
    \end{adjustbox}
    \label{tab:qtype}
    \vspace{-3mm}
\end{table}

\begin{figure}[!t]
\centering
\includegraphics[width=0.3\textwidth]{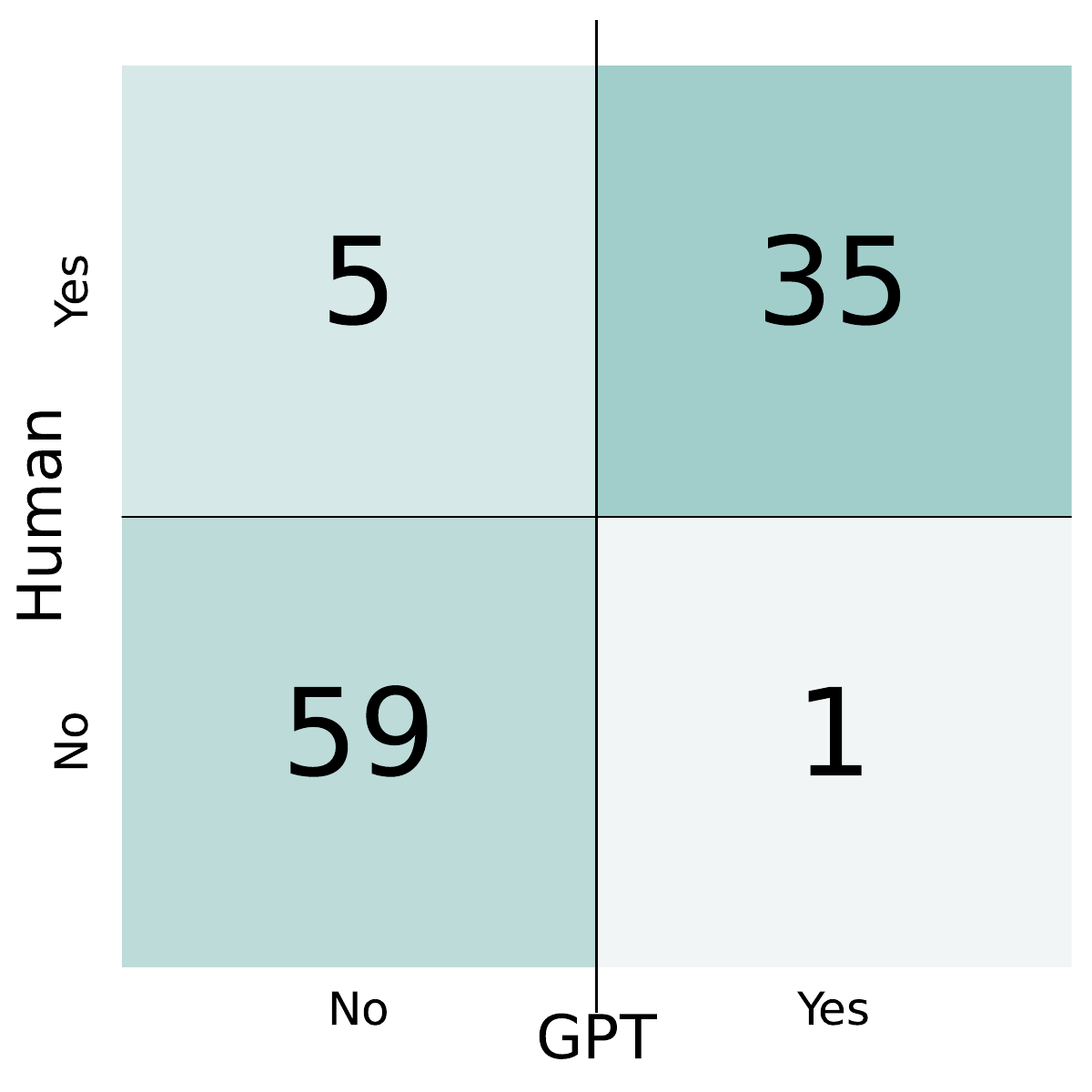}
\caption{Agreement between human and GPT evaluation on 100 instances.}
\label{fig:aggrement}
\vspace{-3mm}
\end{figure}

\begin{table*}[ht!]
  \centering
  \small
  \caption{Performances across all selected models (results based on 30\% data). Dispider performs only MC-QA.}
  \label{tab:all_results}
    \resizebox{1.0\textwidth}{!}{
 \begin{tabular}{l | c c |cc|c| c  c c| c  c| c }
\toprule
\multirow{2}{*}{\textbf{Methods}}
& \multicolumn{5}{c|}{\textbf{Multiple-Choice}} 
& \multicolumn{6}{c}{\textbf{Open-Ended}} \\
\cmidrule(lr){2-6} \cmidrule(lr){7-12}
&  MC-2 & MC-5 & Cur.&Pre.& Avg. 
& Action & Object & Others & Cur. & Pre. & Avg. \\
\midrule
\multicolumn{12}{c}{\textit{Closed-source Models}} \\
\midrule
GPT-5~\cite{openai2025gpt}                       & 64.5 & 54.2 & 57.5 & 55.9 & 56.3    & 50.0        & 44.6        & 43.1        & 51.1         & 44.0          & 46.1         \\
Gemini-2.5 Pro~\cite{comanici2025gemini}     & 60.3&   45.1&   49.9&  47.5&  48.2  & 40.2        & 40.2        & 47.7        & 42.4         & 40.3          & 40.9         \\
\midrule
\multicolumn{12}{c}{\textit{General Open-source Models}} \\
\midrule
        \rowcolor{gray!20}
    \multicolumn{12}{l}{\textit{Parameters > 10B}} \\
    Qwen3-VL-32B-Instruct~\cite{Qwen3-VL}              & 58.3&   35.4&   44.1& 38.4&  40.0 & 41.4        & 36.4        & 37.8        & 40.0         & 37.3          & 38.1         \\
Qwen3-VL-32B-Thinking~\cite{Qwen3-VL}          & 55.2 & 37.9 & 44.3 & 40.0 & 41.3    & 40.1        & 35.5        & 35.8        & 40.6         & 35.4          & 36.9         \\
InternVL3.5-38B-Thinking~\cite{wang2025internvl3}             & 54.6&   41.8&   43.8&  44.6&  44.4  & 35.5        & 34.8        & 37.8        & 35.2         & 35.3          & 35.3         \\
Qwen2.5-VL-32B-Instruct~\cite{bai2025qwen2}          &  49.3&   31.4&   42.9&  31.9&  35.0  & 39.7        & 33.6        & 30.5        & 36.8         & 34.5          & 35.1         \\
InternVL3.5-38B-Instruct~\cite{wang2025internvl3}                & 56.3&   39.0&   41.9&  42.7&  42.5  & 35.0        & 34.1        & 37.8        & 34.3         & 34.9          & 34.7         \\
InternVL3-38B~\cite{zhu2025internvl3}                            & 51.3&   41.9&   44.1&  43.7&  43.8    & 35.7        & 32.5        & 39.1        & 33.3         & 34.5          & 34.1         \\

        \rowcolor{gray!20}
    \multicolumn{12}{l}{\textit{4B <Parameters$\leq$ 10B}} \\
    Qwen2-VL-7B-Instruct~\cite{wang2024qwen2vlenhancingvisionlanguagemodels} & 54.3&   29.6&   36.5&  33.9&  34.6&34.6&36.3&35.1&35.8&35.6&35.7\\

    Qwen3-VL-8B-Instruct~\cite{zhu2025internvl3}        &  55.7&   35.5&   44.4&  37.7&  39.6 & 37.0        & 35.6        & 28.5        & 36.3         & 34.9          & 35.3         \\
    InternVL3-8B~\cite{zhu2025internvl3}                        & 53.6&   38.8&   41.0&  42.1&  41.8  & 31.9        & 36.3        & 36.4        & 35.6         & 34.8          & 35.0         \\
    Qwen2.5-VL-7B-Instruct~\cite{bai2025qwen2}      &  53.3&   32.5&   43.6&  34.0&  36.7   & 34.6        & 35.0        & 31.1        & 37.7         & 33.2          & 34.5         \\
    
LLaVA-Video~\cite{zhang2024video}                  & 52.3&   33.8&   42.7&  35.5&  37.5  & 33.9        & 33.9        & 39.7        & 36.3         & 33.7          & 34.5         \\
Qwen3-VL-8B-Thinking~\cite{Qwen3-VL}&  55.3&   31.2&   40.8&  34.2&  36.0   & 34.1        & 34.1        & 31.8        & 38.4         & 32.0          & 33.9         \\
MiniCPM-V 4.5~\cite{yu2025minicpm}          &50.3&   35.8&   41.2&  37.8&  38.7  & 35.2        & 32.2        & 37.1        & 34.3         & 33.3          & 33.6         \\
LongVU~\cite{shenlongvu}               & 56.0&   31.9&   40.3&  35.3&  36.7  & 31.9        & 33.1        & 36.4        & 30.8         & 34.0          & 33.1         \\
InternVL3.5-8B-Thinking~\cite{wang2025internvl3}                    &  52.0&   36.3&   42.7&  38.2&  39.5    & 31.1        & 33.9        & 34.4        & 33.8         & 32.8          & 33.1         \\
LLaVA-OV-7B~\cite{li2024llava}               &49.7&   32.7&   40.8&  34.3&  36.1   & 31.1        & 32.4        & 38.4        & 33.8         & 32.1          & 32.6         \\
LongVA~\cite{zhang2024longva}&56.3&   28.6&   39.6&  32.1&  34.2  & 33.9        & 31.0        & 29.8        & 34.7         & 30.5          & 31.7         \\
InternVL3.5-8B-Instruct~\cite{wang2025internvl3}                 &  55.6&   35.5&   41.2&  38.9&  39.6  & 32.4        & 26.8        & 27.2        & 31.5         & 27.3          & 28.5         \\
InternVL2.5-8B~\cite{chen2024far}             & 50.3&   35.3&   41.2&  37.2&  38.3  & 27.5        & 21.9        & 19.2        & 27.2         & 21.8          & 23.3         \\
        \rowcolor{gray!20}
    \multicolumn{12}{l}{\textit{Parameters$\leq$ 4B}} \\
    Qwen3-VL-4B-Instruct~\cite{Qwen3-VL}               &  55.7&   35.0&   43.6& 37.4&  39.2&  36.1        & 35.6        & 35.8        & 37.4         & 35.1          & 35.8         \\
    Qwen3-VL-4B-Thinking~\cite{Qwen3-VL}               & 55.3&   29.6&   39.1&  33.1&  34.7&  37.0        & 33.5        & 29.8        & 37.2         & 33.0          & 34.2         \\
    InternVL3.5-4B-Thinking~\cite{zhu2025internvl3}  & 48.3&   36.8&   41.7&  38.1&  39.1&   32.8        & 31.0        & 35.8        & 32.4         & 31.8          & 32.0         \\

InternVL3.5-4B-Instruct~\cite{zhu2025internvl3}      & 49.0&   31.1&   36.7&  34.0&  34.7    & 28.6        & 30.7        & 30.5        & 30.6         & 29.9          & 30.1  \\    
Qwen2.5-VL-3B-Instruct~\cite{bai2025qwen2}            &  54.0&   28.7&   38.8&  31.8&  33.8  & 29.3        & 29.8        & 30.5        & 28.1         & 30.4          & 29.7         \\
  
\midrule
\multicolumn{12}{c}{\textit{Memory-Enhanced Streaming QA Methods}} \\
\midrule
Flash-VStream (Qwen2-VL-7B)~\cite{zhang2025flashvstream} &51.3&  31.1&  36.3& 34.8&  35.2&31.9&29.9	&35.1	&31.5	&30.8&	31.0\\
Dispider (Qwen2-VL-7B)~\cite{qian2025dispider} &45.5&31.2&   31.9& 33.1&  32.7 &-&-&-&-&-&-\\
%Flash-VStream (Qwen2-VL-7B)~\cite{zhang2025flashvstream} & 1\\
\bottomrule
\end{tabular}
}
\end{table*}

\begin{figure*}[!t]
\centering
\includegraphics[width=0.9\textwidth]{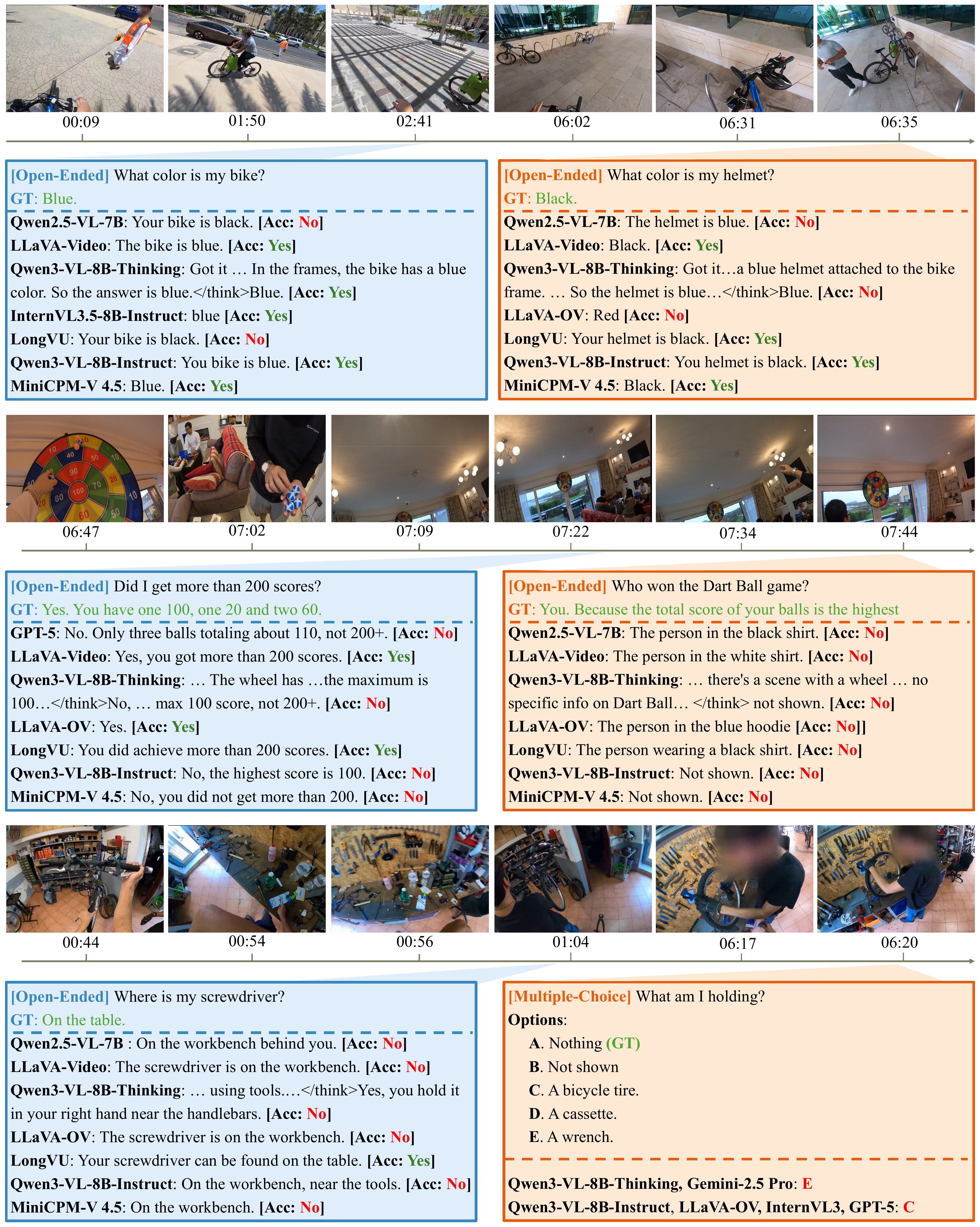}
\caption{Result visualization on~\dataset. Models struggle to correctly answer even simple questions, indicating their severe deficiency in performing ego-grounding in videos to answer personalized questions from the camera wearers. \textbf{Top:} The camera wearer is cycling alongside another man whose bike and helmet differ in color from his. \textbf{Middle:} The camera wearer is playing a dart-ball game with others and achieves the highest score. \textbf{Bottom:} The camera wearer is repairing a bicycle with another man, repeatedly putting down and picking up his screwdriver.}
\label{fig:morecase}
\vspace{-3mm}
\end{figure*}

\begin{table*}[h!] % The [h!] suggests LaTeX to place the table *here* if possible
  \centering % Center the box within the page width
    \caption{Prompts for Gemini-2.5 Pro to generate distracting answers in multi-choice QA.} % This adds the caption
    \vspace{-3mm}
  \begin{promptbox}
    You are an expert in egocentric video analysis and a creative multiple-choice question designer. Your task is to generate a complete set of multiple-choice options for each question, which includes refining the correct answer and creating four plausible distractors.

**IMPORTANT INSTRUCTIONS:**

1. Analyze the Video: Carefully analyze the provided video. The timestamps ``question\_moment'' and ``answer\_moment'' are critical.

2.  Create Multiple-Choice Options: For each question, you will generate a single list of options. The **very first option** in your list MUST be the refined correct answer, followed by exactly four distractors.

3.  Generate Distractors: Following the refined answer, create four distractors. Each should be formatted as `"distractor (Criterion)"` and ideally meet one of the following criteria:

    \begin{itemize}
        \item[] **Criterion A (Contextual Confusion)**: The item is present and visible during the `answer\_moment` but is not the correct answer.
    
    \item[] **Criterion B (Environmental Objects)**: The item is present in the video but is not being used or worn by the person recording.
    
    \item[] **Criterion C (Temporal Misdirection)**: The item appears *after* the `question\_moment` but is not the correct answer for that specific question.
    
    \item[] **Criterion D (Attentional Decoy)**: The item is present and visible during the `question\_moment` to distract the viewer's attention.
    \end{itemize}
    Additionally, the phrasing of the distractors should be as similar as possible to the correct answer.
    
4.  Fallback Rule: If a distractor cannot meet the above criteria, it must at least be an object or action present in the video and be a plausible, yet incorrect, answer.

5.  Special Case for "Yes/No": If the answer is purely "yes" or "no" (i.e., without any additional explanation), you MUST return a list with **only two options**: the correct answer formatted as "Yes (Refined Answer)" or "No (Refined Answer)", and the opposite option formatted as "No (Distractor)" or "Yes (Distractor)". Otherwise, you should return distractor answers with explanation.

6.  Input Format: You will receive the list of question-answer pairs in the following JSON format:

        question: \{QUESTION\}\\
        answer: \{ANSWER\}\\
        question\_moment: \{QUESTION\_MOMENT\}\\
        answer\_moment: \{ANSWER\_MOMENT\}

7.  REQUIRED Output Format: You MUST return your response as a single, valid JSON array (a list of lists). The first item in each inner list must be the refined answer.

    **Example Output:**

    [

    \begin{itemize}
        \item[] "ground truth (Refined Answer)",

        \item[]    "distractor A (Contextual Confusion)",
         \item[]   "distractor B (Environmental Objects)",
         \item[]   "distractor C (Temporal Misdirection)",
          \item[]  "distractor D (Attentional Decoy)"
    \end{itemize}

    ]

Please now generate the refined answers and distractors for the provided video and QA pairs.
  \end{promptbox}
  \label{tab:distractor_gen} % This is the label for referencing
\end{table*}

\section{Evaluation Details}
\textbf{Prompt Details}. \cref{tab:gen_prompt} shows the prompt we used to test the model performances. Following the official inference code, we add a thinking system prompt for InternVL3.5-Thinking series~\cite{wang2025internvl3}, and insert a time instruction right after the video frames for LLaVA-Video~\cite{zhang2024video}.\\

\begin{table*}[!t]
    \small
    \centering
  \caption{Prompts for models to answer questions.}
  \label{tab:gen_prompt}
  \begin{tabularx}{\textwidth}{c|X}
    \hline
    \textbf{Task}&\textbf{General Prompt}\centering\arraybackslash \\
    \hline
   Open-ended & You are a first-person AI assistant integrated into a head-mounted camera. Your primary mission is to answer questions from the user (the camera wearer) about their own actions, objects, and environment as seen through your lens. You should answer directly to the user. The question is asked at the moment of the last frame in the video. Key Instructions: 1. First-Person Context: All references to 'I', 'me', or 'my' are about the user. You must distinguish between the user's actions/objects and those of other people visible in the video. 2. Grounding: Base your answers primarily on the visual evidence in the video. If the information is not present or cannot be reasonably inferred from the video, state that it is not shown. 3. For any question that requires a 'Yes' or 'No' response, you MUST follow it with a brief explanation for your reasoning. 4. All responses must be direct and clear. \{VIDEO\_CONTENT\}. Question: \{QUESTION\}.\\
   \hline
   Multiple-choice & You are a first-person AI assistant integrated into a head-mounted camera. Your primary mission is to answer questions from the user (the camera wearer) about their own actions, objects, and environment as seen through your lens. You should answer directly to the user. The question is asked at the moment of the last frame in the video. Key Instructions: 1. All references to 'I', 'me', or 'my' are about the user. You must distinguish between the user's actions/objects and those of other people visible in the video. 2. Base your answers primarily on the visual evidence in the video.  \{VIDEO\_CONTENT\}. Question: \{QUESTION\}. Options: \{2/5 OPTIONS\}. There is only one correct option. Please only response with the letter of the correct option.\\
   \hline
   Removed personalized cues & You are a first-person AI assistant integrated into a head-mounted camera. Your primary mission is to answer questions from the user (the camera wearer) about their own actions, objects, and environment. You should answer directly to the user. The question is asked at the moment of the last frame in the video. Key Instructions: 1. Grounding: Base your answers primarily on the visual evidence in the video. If the information is not present or cannot be reasonably inferred from the video, state that it is not shown. 2. For any question that requires a 'Yes' or 'No' response, you MUST follow it with a brief explanation for your reasoning. 3. All responses must be direct and clear. \{VIDEO\_CONTENT\}. Question: \{QUESTION\}.\\
   \hline
  \end{tabularx}
\end{table*}

\noindent\textbf{Automatic Evaluation}. We feed the prompt in \cref{tab:oe_eval} to GPT-5 mini~\cite{openai2025gpt} to evaluate whether the responses from a tested model (\eg, \gemini) match the ground-truth answers. After two rounds of manual review and refinement of the evaluation prompt, we achieve an agreement rate of 94\% with human judgment on 100 instances (\cref{fig:aggrement}).

\begin{table*}[h!] % The [h!] suggests LaTeX to place the table *here* if possible
  \centering % Center the box within the page width
  \caption{Prompts for GPT-5 mini to evaluate open-ended answers.} % This adds the caption
  \vspace{-3mm}
  \begin{promptbox}
   You are an intelligent chatbot designed for evaluating the correctness of generative outputs for question-answer pairs. Your task is to compare the predicted answer with the correct answer and determine if they match meaningfully. Here's how you can accomplish the task:

   ------ ------ ------

   \#\#INSTRUCTIONS:

   - Focus on meaningful matches: Assess whether the predicted answer and the correct answer have a meaningful match, not just literal word-for-word matches.

    - Criteria for Correctness: The predicted answer is considered correct if it reasonably matches the standard answer, recognizing that synonyms or varied expressions that convey the same or similar meaning are acceptable.

    - If the predicted answer's yes/no conclusion conflicts with the correct answer, it is incorrect.

    - The predicted answer is considered correct if it contains the core descriptive information and does not contradict the correct answer, even if some non-critical details are missing.

    - Flexibility in Evaluation: Use judgment to decide if variations in the predicted answer still correctly address the question, even if they do not directly replicate the correct answer's phrasing.

    \qquad Please evaluate the following question-answer pair:

    \qquad Question: \{QUESTION\}
    Correct Answer: \{GT-ANSWER\}
    Predicted Answer: \{MODEL-RESPONSE\}

    Provide your evaluation result only as a yes/no and score where the score is an integer value between 0 and 5, with 5 indicating the highest meaningful match.

    Please generate the response in the form of a valid JSON string with keys 'pred' and 'score'. 
    
    \qquad For example: \{"pred": "yes", "score": 5\}, \{"pred": "no", "score": 1\}.
  \end{promptbox}
  \label{tab:oe_eval} % This is the label for referencing
  \vspace{-3mm}
\end{table*}

\section{More Experiment Analyses}
\cref{tab:all_results} reports the performance of all selected models on~\dataset. For efficiency, we evaluate on a randomly sampled 30\% subset (1,500 instances). We further include two recent memory models Dispider~\cite{qian2025dispider} and Flash-VStream~\cite{zhang2025flashvstream} that are specialized for long streaming VideoQA for comparison. 

By comparing among models of different parameter scales (\eg, 4B $\rightarrow$ 8B $\rightarrow$ 32B $\rightarrow$ 38B), we find that larger models do not consistently outperform the smaller ones.
Even within the same model family, smaller variants (\eg, 4B) can achieve competitive or even superior performance, as seen in the InternVL3 series. Additionally, the two long Video-LLMs, LongVA~\cite{zhang2024long} and LongVU~\cite{shenlongvu}, do not exhibit outstanding results despite processing substantially more frames (128 \vs 32 of other general models). Similarly, the two long streaming QA methods also fail to achieve their desired level of performance. Surprisingly, the relatively old model Qwen2-VL leads the models of 7B sizes in open-ended QA, surpassing many recent systems.
The above analyses collectively suggest that the explored problem of ego-grounding and personalized understanding, despite with significant practical value for personal assistance, is largely overlooked in existing technique evolution, and highlight the importance of our benchmark and analyses towards advancements in these fields.

By further analyzing performance across different question categories. We find that most models perform better on ``Action'' (\vs ``Object'') questions. This indicates that identifying ``what I am doing'' is easier than disambiguating ``my thing'' from those of others in egocentric videos. The findings may depart from our common understanding about video object and action recognition. We visualize some examples in \cref{fig:morecase} for better understanding of results. Moreover, by comparing between ``Current'' and ``Previous'' questions, we observe a clear performance gap of almost all models: They perform better in answering questions whose answers are visible at the question moments. This suggests that the models are capable of better ground the camera wears and their belongings in the question moments, but such strength drops sharply when the answers are out of scene.


\end{document}